\documentclass[lettersize,journal]{IEEEtran}
\usepackage{amsmath,amsfonts}
\usepackage{algorithmic}
\usepackage{algorithm}
\usepackage{array}
\usepackage[caption=false,font=normalsize,labelfont=sf,textfont=sf]{subfig}
\usepackage{textcomp}
\usepackage{stfloats}
\usepackage{url}
\usepackage{verbatim}
\usepackage{graphicx}
\usepackage{cite}
\usepackage{xcolor}
\usepackage{color, colortbl}
\usepackage{multirow}
\usepackage{booktabs}
\usepackage{cleveref}

\newcommand{\eg}{{\emph{e.g.}}}

\begin{document}

\title{TiGDistill-BEV: Multi-view BEV 3D Object Detection via Target Inner-Geometry Learning Distillation}

\author{Shaoqing Xu, Fang Li, Peixiang Huang, Ziying Song,  Zhi-Xin Yang, \textit{Member, IEEE}
\thanks{This work was funded in part by the Science and Technology Development Fund, Macau SAR (Grant no. 0018/2019/AKP, SKL-IOTSC(UM)-2021-2023 and 0059/2021/AFJ), in part by the Guangdong Science and Technology Department, China (Grant no. 2020B1515130001), in part by the University of Macau (Grant No.: MYRG2020-00253-FST and MYRG2022-00059-FST), and in part by the Zhuhai UM Research Institute (Grant No.: HF-011-2021). \emph{(Corresponding author: Zhi-Xin Yang).}
}
\thanks{Shaoqing Xu and Zhi-Xin Yang are with the State Key Laboratory of Internet of Things for Smart City and Department of Electromechanical Engineering, University of Macau, Macau 999078, China (e-mail: shaoqing.xu@connect.um.edu.mo, zxyang@um.edu.mo)
}

\thanks{Fang Li is with School of Mechanical Engineering, Beijing Institute of Technology. (e-mail:a319457899@163.com)}

\thanks{Peixiang Huang is with the College of Engineering, Peking University, Beijing, China (e-mail: huangpx@stu.pku.edu.cn)
}
\thanks{Ziying Song is with School of Computer and Information Technology, Beijing Key Lab of Traffic Data Analysis and Mining, Beijing Jiaotong University, Beijing 100044, China (e-mail: 22110110@bjtu.edu.cn)
}

}


\maketitle

\begin{abstract}
Accurate multi-view 3D object detection is essential for applications such as autonomous driving. Researchers have consistently aimed to leverage LiDAR’s precise spatial information to enhance camera-based detectors through methods like depth supervision and bird-eye-view (BEV) feature distillation. However, existing approaches often face challenges due to the inherent differences between LiDAR and camera data representations.
In this paper, we introduce the \textbf{TiGDistill-BEV}, a novel approach that effectively bridges this gap by leveraging the strengths of both sensors. Our method distills knowledge from diverse modalities(\eg, LiDAR) as the teacher model to a camera-based student detector, utilizing the Target Inner-Geometry learning scheme to enhance camera-based BEV detectors through both depth and BEV features by leveraging diverse modalities. Specially, we propose two key modules: an inner-depth supervision module to learn the low-level relative depth relations within objects which equips detectors with a deeper understanding of object-level spatial structures, and an inner-feature BEV distillation module to transfer high-level semantics of different keypoints within \textbf{foreground targets}. To further alleviate the domain gap, we incorporate both inter-channel and inter-keypoint distillation to model feature similarity. Extensive experiments on the nuScenes benchmark demonstrate that TiGDistill-BEV significantly boosts camera-based only detectors
achieving a state-of-the-art with 62.8\% NDS and surpassing previous methods by a significant margin. The codes is available at: \textcolor{magenta}{https://github.com/Public-BOTs/TiGDistill-BEV.git}
\end{abstract}

\begin{IEEEkeywords}
3D Object Detection, Knowledge Distillation, BEV
\end{IEEEkeywords}

\section{Introduction}
\label{sec:intro}

\begin{figure}[t!] 
\centering
\includegraphics[width=1\linewidth]{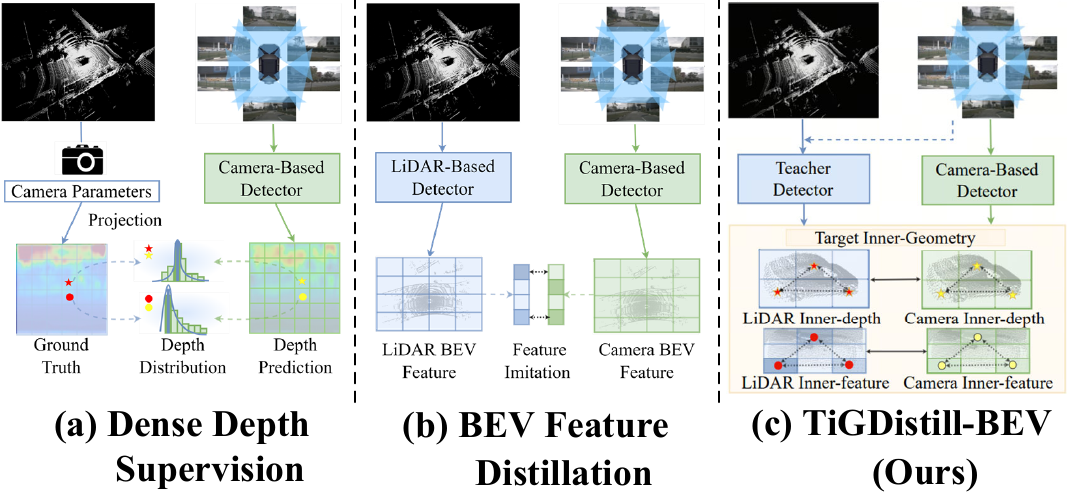} 
\caption{\textit{Different LiDAR-to-Camera Learning Schemes}: (a) Dense Depth Supervision, which directly supervises the categorial depth distribution of every valid pixel in the whole depth map, \textbf{(b)} BEV Feature Distillation, which constrainedly aligns the value of BEV feature between different modalities, \textbf{(c)} Our Target Inner-Geometry Learning, which utilizes both the low-level inner-depth relations and the high-level inner-feature semantics of foreground targets.} 
\label{fig:teaser_figure}
\end{figure}

\IEEEPARstart{M}{ulti-view} 3D object detection plays a crucial role in enabling intelligent systems to perceive their surroundings accurately, which has made remarkable strides in various applications. LiDAR-based methods\cite{yin2021center,song2024robofusion,li2023pillarnext,tang2020improving} excel due to their inherent capture of rich spatial structures. In contrast, camera-based methods\cite{yang2023mix,song2022wsamf,song2024voxelnextfusion,jiang2024far3d,tao2023pseudo}, while cost-effective, lack direct access to geometric depth information. To address these performance disparities, existing multi-modal 3D object detection leverages LiDAR’s spatial cues to enhance camera-based detectors\cite{yu2023pipc,song2024graphbev,xu2021fusionpainting,song2023graphalign++,xu2024multi}. This enhancement primarily falls into two schemes, as illustrated in (a) and (b) in the Figure.\ref{fig:teaser_figure}. Dense Depth Supervision, \eg, BEVDepth~\cite{b7}. These methods project the input LiDAR points onto image planes as depth maps, and explicitly supervise the categorical depth prediction within both foreground and background regions. BEV Feature Distillation employing the teacher-student paradigm, \eg, BEVDistill \cite{b9}, compels the camera-based detector to imitate the BEV representation of a pre-trained LiDAR-based detector (the teacher).
The student inherits the encoded high-level BEV semantics from the teacher by directly mimicking the BEV features.

However, existing methods fall short in capturing the inner-geometric characteristics of foreground targets. Inner geometry of an object includes its low-level spatial contours and high-level part-wise semantic relations, which are significant for precise object recognition and localization.
For instance, BEVDepth simply employs pixel-level depth supervision without considering relative depth within objects, and BEVDistill applies foreground-guided distillation but neglects the inner relations of BEV features. Moreover, methods~\cite{b52,ma2024licrocc}, the direct channel-level alignment enforced in BEV feature distillation can be detrimental due to the modality gap between camera and LiDAR BEV features. To alleviate this issue, we propose \textbf{TiGDistill-BEV}, a novel multi-modality distillation learning scheme that
integrates the inner-geometry of foreground targets into camera-based detectors for multi-view BEV 3D object detection. As shown in Figure. \ref{fig:teaser_figure} (c), we simultaneously perform target inner-geometry learning for both depth prediction and BEV representation learning.

\begin{figure}[t!]
\centering
        \includegraphics[width=\columnwidth]{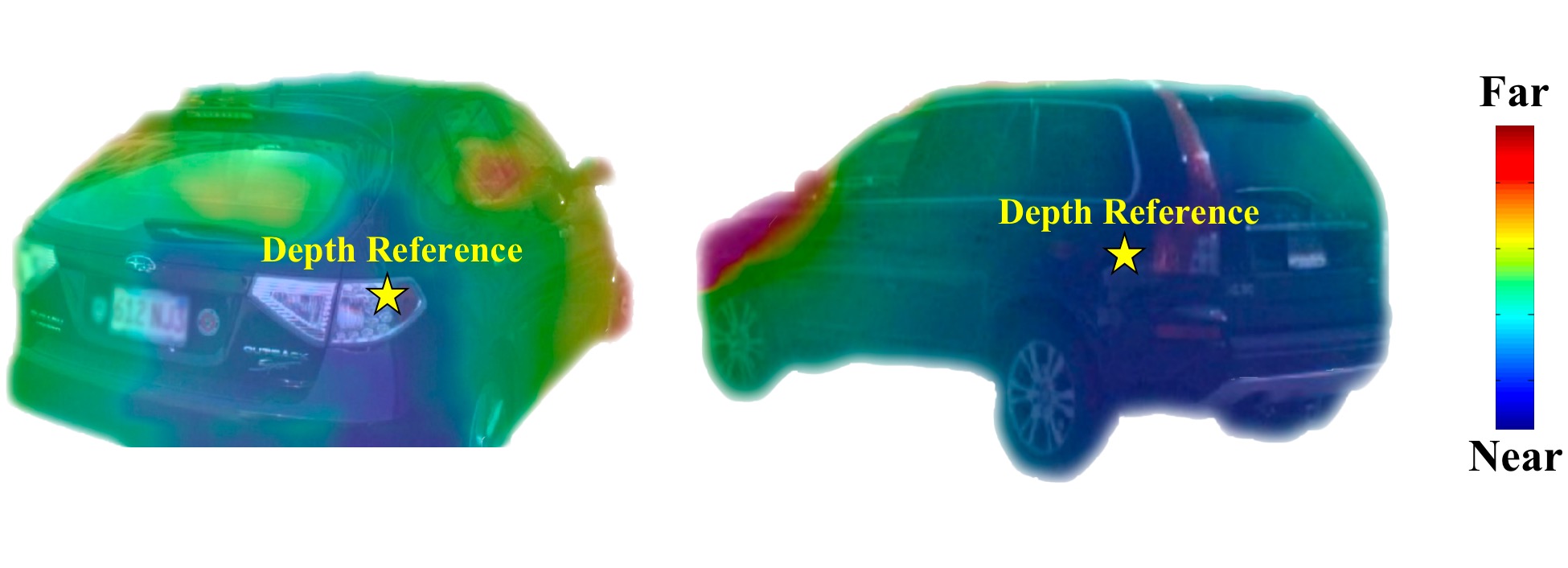}
        \caption{\textit{Inner-depth Supervision.} We guide the camera-based detector to learn the relative spatial structures within the target foreground areas. A depth reference point (dotted in yellow) is adaptively selected to calculate relative depth.}
        \label{fig:fig2}
\end{figure}

\begin{figure}[t!]
\centering
        \includegraphics[width=.9\columnwidth]{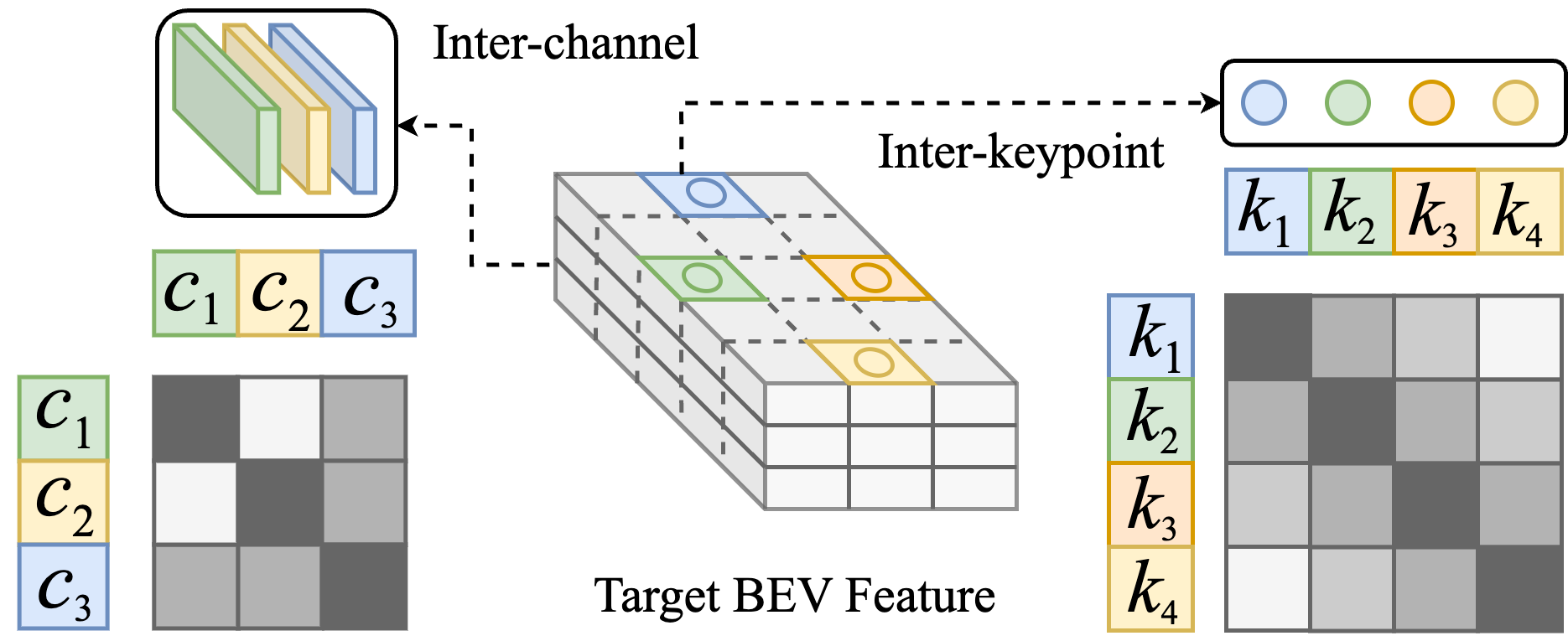}
        \caption{\textit{Inner-feature BEV Distillation.} We conduct inter-channel and inter-keypoint feature distillation in BEV space for the camera-based detector,   which alleviates the cross-modal semantic gap and boosts inner-geometry learning.}
        \label{fig:fig3}
\end{figure}

Firstly, besides the absolute depth map prediction~\cite{b7,b51}, we introduce an inner-depth supervision module within pixels of different foreground targets. A reference point of depth is adaptively selected for each target to obtain the relative depth relationships shown in Figure. \ref{fig:fig2}, which contributes to high-quality depth map prediction with better target structural understanding. Secondly, we propose an inner-feature distillation module, which imitates the high-level foreground BEV semantics generated by a pre-trained multi-modal detector. Different from previous methods of dense and strict feature distillation~\cite{b9,b52}, we adaptively sample several keypoints within each BEV foreground area and guide the camera-based detector to learn their inner feature-similarities shown in Figure. \ref{fig:fig3}, which are in both inter-channel and inter-keypoint manners. 
This approach enables the camera-based detector to not only inherit high-level, part-wise LiDAR semantics but also relieve the modality gap by avoiding strict feature mimicry.
Our extensive experiments consistently demonstrate performance improvements compared to the baseline models.

Contributions of TiGDistill-BEV are summarized below:
\frenchspacing
\begin{itemize}
    \item We introduce an inner-depth supervision module that enables us to capture the internal depth relations between different parts of each foreground target and leads to a better target depth map prediction which is important to the perspective transformation to obtain the BEV feature.
    \item We propose an inner-feature BEV distillation module to transfer the well-learned knowledge from diverse modalities to camera-based BEV representations with the \textbf{inner-geometry learning} for high-level BEV semantics instead of directly feature alignment.
    \item Extensive experiments have confirmed our effectiveness to enhance the  \textbf{camera-based} multi-view BEV 3D object detection. On nuScenes val set, the powerful BEVDepth is boosted by \textbf{+2.3\%} NDS and \textbf{+2.4\%} mAP, and with further enhancements of \textbf{+3.9\%} NDS and \textbf{+4.8\%} mAP on test set.
\end{itemize}

\begin{figure*}[t!]
  \centering
  \includegraphics[width=0.9\textwidth]{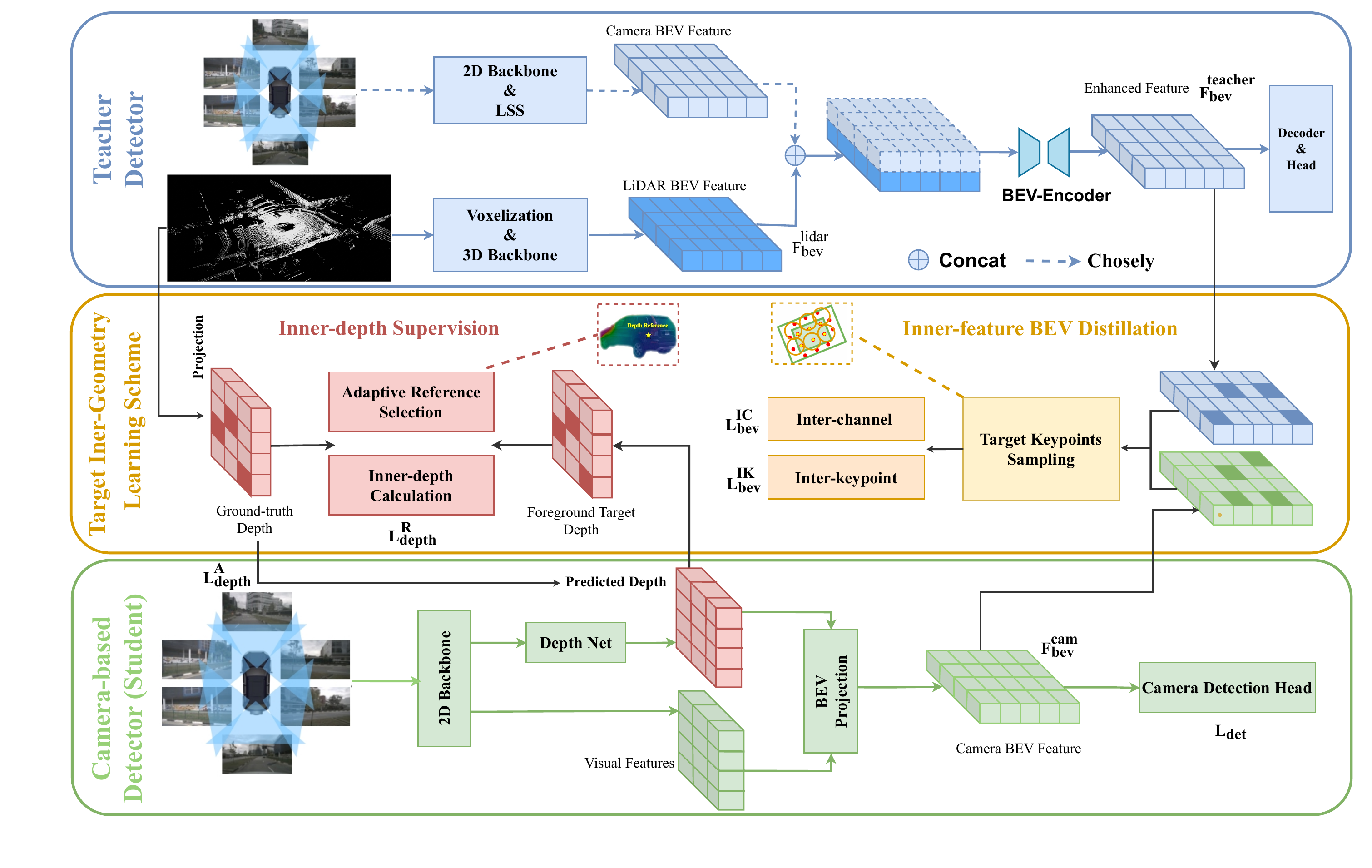}
  \caption{\textit{Overall Framework of TiGDistill-BEV,} which contains a pre-trained teacher model, a camera-based detector as student, and a target inner-geometry scheme for cross-modal learning. Our proposed learning paradigm bridges the modalities gap by transferring the inner-geometry semantics from the teacher modality via two components, an inner-depth supervision for foreground relative depth, and an inner-feature BEV distillation from both channel-wise and keypoint-wise.}
  \label{fig:framework}
\end{figure*}

\section{Related Work}
\subsubsection{Camera-based 3D Object Detection} 
Camera-based 3D object detection has been widely used for applications like autonomous driving since its low cost. 
Bird’s-Eye-View~(BEV), as a unified representation of surrounding views, has attracted much attention. BEVDet series \cite{b19,b23} utilize LSS \cite{b20} to transform 2D image features into 3D BEV representation. BEVDepth leverages the accurate depth estimation for BEV 3D object detection supervised by projected LiDAR points. BEVFormer \cite{b11} automates the camera-to-BEV process with learnable attention modules and BEV queries in 3D space. Following query-based 2D object detectors, DETR3D \cite{b13} and PETR series\cite{b21,b24} adopt 3D object queries and integrate 3D reference points for feature aggregation. StreamPETR \cite{wang2023exploring} and Sparse4D \cite{lin2022sparse4d,lin2023sparse4d} achieve substantial improvements using object-centric temporal modeling paradigm.
 

\subsubsection{Depth Estimation} Depth estimation is a classical problem in computer vision. These method can be divided into single-view depth estimation and multi-view depth estimation. Single-view depth estimation is either regarded as a regression problem of a dense depth map or a classification problem of the depth distribution. \cite{b26, b27, b28, b29, b30} generally build an encoder-decoder architecture to regress the depth map from contextual features. Multi-view depth estimation methods usually construct a cost volume to regress disparities based on photometric consistency \cite{b31, b32, b33, b34, b35, b62}. For 3D object detection, previous methods~\cite{b61,b51,b52} also introduce additional networks for depth estimation to improve the localization accuracy in 3D space.
Notably, MonoDETR \cite{b47,b48} proposes to only predict the foreground depth maps instead of the dense depth values, but cannot leverage the advanced geometries provided by LiDAR modality. Different from them, our TiGDistill-BEV conducts inner-depth supervision that captures local sptial structures of different foreground targets.

\subsubsection{Relationship Supervision} Some related works have investigated channel-wise and pixel-wise relationship supervision in various domains. \cite{b66} studied pixel-wise relationships in image style transfer and found that matching higher layer style representations preserves local image structures at a larger scale, resulting in smoother visuals. \cite{b67} proposed similarity-preserving knowledge distillation, guiding the student network towards teacher network's activation correlations. If two inputs produce similar activations in the teacher network, the student network should be guided towards a similar configuration. \cite{b68} introduced a channel-wise relationship preserving loss for visualizing adapted knowledge in domain transfer. They claimed that channel-wise relationships remain effective after global pooling, unlike pixel-wise relationships, which can be overshadowed pre-classifier. Our TiGDistill-BEV has also been inspired by these works and explored the designs of learning the internal relationships of foreground objects with diverse modalities.

\subsubsection{Knowledge Distillation} Knowledge distillation has shown very promising ability in transferring learned representation from the larger model (teacher) to the smaller one (student). Prior works \cite{b37, b38, b39, b40} are proposed to help the student network learn the structural representation for better generalization ability. Such teacher-student paradigms have also been extended to other vision tasks, including action recognition \cite{b41}, video caption \cite{b42}, 3D representation learning~\cite{b59,b49,b50,b60}, object detection \cite{b43, b44,b65} and semantic segmentation \cite{b45, b46}. However, only a few of works consider the multi-modal setting between different sensor sources. For 3D representation learning, there are some interesting approaches. I2P-MAE~\cite{b49} leverages Masked Autoencoders to distill 2D pre-trained knowledge into 3D transformers. BEV-LGKD \cite{b10} generates the foreground mask and view-dependent mask for better localization. BEVDistill \cite{b9} transfer knowledge from LiDAR feature to the cam feature by dense feature distillation and sparse instance distillation. UniDistill \cite{b65} focuses on transferring knowledge from multi-modality detectors to single-modality detectors in a universal manner. X$^3$KD \cite{b69} is a knowledge distillation framework for multi-camera 3D object detection, leveraging cross-modal and cross-task information by distilling knowledge from LiDAR-based detectors and instance segmentation teachers. DistillBEV \cite{b70} involves feature imitation and attention imitation losses across multiple scales, enhancing feature alignment between a LiDAR-based teacher and a multi-camera BEV-based student detector. BEVSimDet \cite{b71} proposes a simulated multi-modal student to simulate multi-modal features with image-only input. VCD \cite{b72} presents some useful designs for temporal fusion and introduce a fine-grained trajectory-based distillation module. FD3D \cite{b73} uses queries for masked feature generation and then intensify feature representation for refined distillation. 

Our TiGDistill-BEV also follows such teacher-student paradigm and effectively distills knowledge from the diverse modalities into the camera modality. By modeling the relative relationships inside foreground objects in a novel way, the performances of camera-only detectors are further enhanced.


\section{Method}
\label{sec:method}
The overall architecture of TiGDistill-BEV is illustrated in Figure. \ref{fig:framework}. We begin by detailing the baseline models employed in \cref{sec:Baseline Models}. Following this, the proposed target inner-geometry learning scheme is introduced in \cref{sec:Inner-depth Supervision} and \cref{sec:Inner-feature BEV Distillation}, which elaborate on how  TiGDistill-BEV distills the inner-geometry characteristics through inner-depth supervision and inner-BEV feature distillation, respectively.  Finally, the overall loss function of our framework is presented in \cref{sec:overall_loss}.

\subsection{\textbf{Baseline Models}}
\label{sec:Baseline Models}

\subsubsection{\textbf{Student Camera-based Detector}}
We adopt BEVDepth as our student model which takes multi-view images to extract the $C$-channel visual features $\{F_i\}_{i=1}^6$, where $F_i \in {\mathbb{R}^{C\times H_v\times W_v}}$, $i$ and $H_v, W_v$ denote the number of camera input and size of feature maps. These features are fed into a shared depth network to generate the categorical depth map, $\{D_i\}_{i=1}^6$, where ${D_i}\in {\mathbb{R}^{K\times H_v\times W_v}}$, and {K} denotes the pre-defined number of depth bins. During the training, BEVDepth adopts dense depth supervision to predicted depth maps, which projects the paired LiDAR input onto multi-view image planes to construct pixel-by-pixel absolute depth ground truth, $\{D_i^{gt}\}_{i=1}^6$, where ${D_i^{gt}}\in {\mathbb{R}^{1\times H_v\times W_v}}$. Following~\cite{b20}, the multi-view features are projected into a unified BEV representation via the predicted depth maps, and further encoded by a BEV encoder, denoted as $F^{cam}_{\rm bev}\in {\mathbb{R}^{C\times H_{\rm bev}\times W_{\rm bev}}}$. Finally, the detection heads operate on this to predict 3D objects. We represent the two basic losses of the student model as $\mathcal{L}_{\rm{depth}}^{A}$ and $\mathcal{L}_{\rm{det}}$, denoting the Binary Cross Entropy loss for dense absolute depth values and the 3D detection loss, respectively.

\subsubsection{\textbf{Teacher Detector}}
We investigate the effect of different teacher models for knowledge distillation. Here, we adopt {Centerpoint\cite{yin2021center}, PillarNeXt\cite{li2023pillarnext} and BEVFusion\cite{liu2023bevfusion}} as teacher models which represent the LiDAR and camera-LiDAR fusion-based detectors. We obtain the $C$-channel teacher BEV feature $F^{teacher}_{\rm bev}\in {\mathbb{R}^{C\times H_{\rm bev}\times W_{\rm bev}}}$ through encoder modules, which has the same feature size as $F^{cam}_{\rm bev}$ from the student detector. While $F^{teacher}_{bev}$ can provide sufficient geometric and semantic knowledge for student BEV feature when the teacher model has been well pre-trained, especially in the target foreground areas. Note that the teacher model is merely required during training for cross-modal learning.

\begin{figure}[h]
\centering
    \includegraphics[width=\columnwidth]{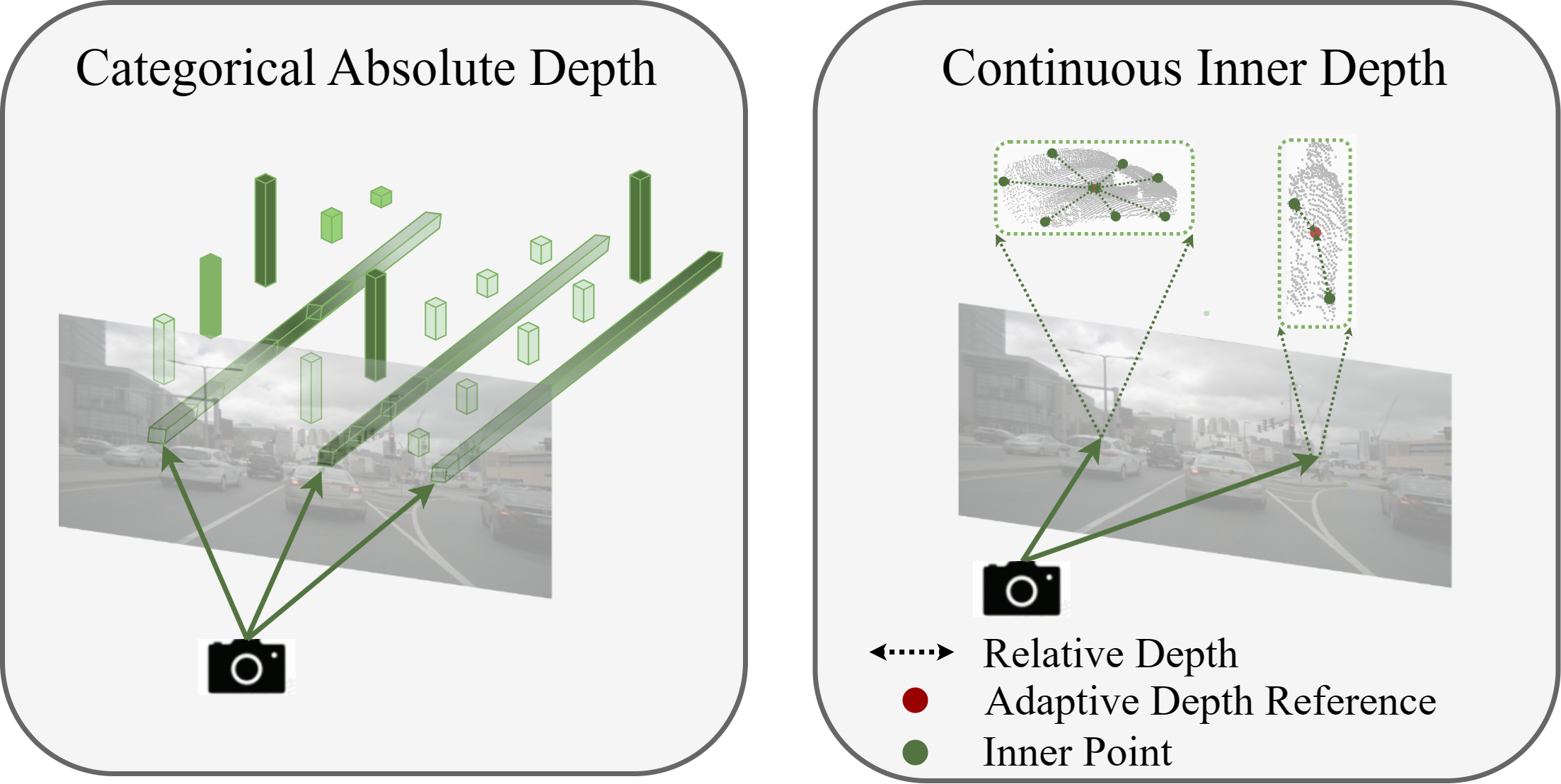}
    \caption{\textit{Comparison Categorical Absolute Depth and Continuous Inner Depth.} Employing the inner-depth supervision with continuous depth values to guide camera-based student to learn local spatial structures of foreground object targets.
    }
    \label{fig:relative_depth}
\end{figure}

\subsection{\textbf{Inner-depth Supervision}}
\label{sec:Inner-depth Supervision}

In addition to the dense absolute depth supervision, we propose to guide the student model to learn the inner-depth geometries in different target foreground areas. As illustrated in Figure. \ref{fig:relative_depth}, for the instance level, the existing absolute depth supervision with categorical representation ignores the relative structural information inside each object and provides no explicit fine-grained depth signals. Therefore, we propose to additionally conduct inner-depth supervision with continuous values from the LiDAR projected depth maps, which effectively enhances the network’s ability to capture the inner-geometry of object targets.

\subsubsection{\textbf{Foreground Target Localization}}
To accurately obtain the inner-depth values, we first localize the foreground pixels for each object target in the depth maps by projecting the corresponding 3D LiDAR points inside the box onto different image planes. This process yields the pixels within foreground object areas on both the predicted and ground-truth depth maps, $\{D_i, D_i^{gt}\}_{i=1}^6$. These foreground pixels roughly depict the geometric contour of different target objects and well facilitate the subsequent inner-depth learning. We take the $i$-th view as an example and omit the index $i$ in the following texts for simplicity. Suppose there exist $M$ target objects on the image, we denote the foreground depth-value set for the $M$ objects as $\{S_j, S_j^{gt}\}_{j=1}^M$, where each $\{S_j, S_j^{gt}\}$ includes the foreground categorical depth prediction and ground-truth depth values for the $j$-th target.

\subsubsection{\textbf{Continuous Depth Representation}}
Unlike the categorical representation used for absolute depth, we
employ continuous values to represent the predicted inner depth of foreground targets, which reflects more fine-grained geometric variations. For pixel $(x, y)$ of the $j$-th target object $S_j$, the predicted possibility of $k$-th depth bin is denoted as $S_j(x, y)[k]$, where $1\le k\le D$. Then, we calculate the continuous depth value $\hat{S_j}(x, y)$ for the pixel $(x, y)$ as follows:
 \begin{equation}
 \label{eq:FM}
 \begin{aligned}
    \hat{S_j}(x, y) = {\sum_{k=1}^D({\hat{S}[k]\cdot S_j(x, y)[k]})},
 \end{aligned}
 \end{equation}
where $\hat{S}[k]$ denotes the depth value of the $k$-th bin center. By this, we convert the categorical depth prediction for each target object, $\{S_j\}_{j=1}^M$, into a continuous representation, denoted as $\{\hat{S_j}\}_{j=1}^M$.

\subsubsection{\textbf{Adaptive Depth Reference}}
 We propose an adaptive depth reference for different foreground targets to compute the relative depth values. Specifically, based on the the predicted continuous depth values in $\{\hat{S}_j\}_{j=1}^M$, we select the pixel with the \textit{smallest depth} prediction error as the reference point for each target, and correspondingly set its depth value as the depth reference, as shown in Figure. \ref{fig:relative_depth}. For the $j$-th target with the ground-truth inner-depth $\{\hat{S}_j, \hat{S}^{gt}_j\}_{j=1}^M$, we calculate the depth reference point $(x_r, y_r)$ by
  \begin{equation}
 \label{eq:FM}
 \begin{aligned}
    (x_r, y_r) = \mathop{\text{Argmin}}_{(x, y)\in \hat{S}_j} \left ({\hat{S}^{gt}_j}(x, y) - {\hat{S}_j}(x, y)\right ).
 \end{aligned}
 \end{equation}
 Then, the predicted and ground-truth reference depth values are denoted as $\hat{S_j}(x_r, y_r)$ and $\hat{S}^{gt}_j(x_r, y_r)$, respectively. By adaptively selecting the reference point with the smallest error, the inner-depth distribution can dynamically adapt to objects with different shapes and appearances, which stabilizes the learning for some truncated and occluded objects.

\subsubsection{\textbf{Inner-depth Calculation}}
Using the reference depth value, we compute the relative depth values within each target
object’s foreground area. For pixel $(x, y)$ of the $j$-th target $\{\hat{S_j}, \hat{S}_j^{gt}\}$, the predicted and ground-truth inner-depth values are denoted as follows: 
\begin{equation}
 \label{eq:FM}
 \begin{aligned}
    r\hat{S_j}(x, y) &= \hat{S_j}(x, y) - \hat{S_j}(x_r, y_r),\\
    r\hat{S}^{gt}_j(x, y) &= \hat{S}^{gt}_j(x, y) - \hat{S}^{gt}_j(x_r, y_r).
 \end{aligned}
 \end{equation}
 We denote the obtained predicted and ground-truth inner-depth value sets for $M$ target objects as $\{\hat{R_j}, \hat{R}_j^{gt}\}_{j=1}^M$. Finally, we supervise the inner-depth prediction by an L2 loss, formulated as follows:
 \begin{equation}
\label{eq:FM}
    \mathcal{L}_{\rm{depth}}^{R} = \sum_{j=1}^M ||\hat{R}_{j}-\hat{R}^{gt}_{j}||_2.
\end{equation}


\begin{figure}[h]
\centering
    \includegraphics[width=\columnwidth]{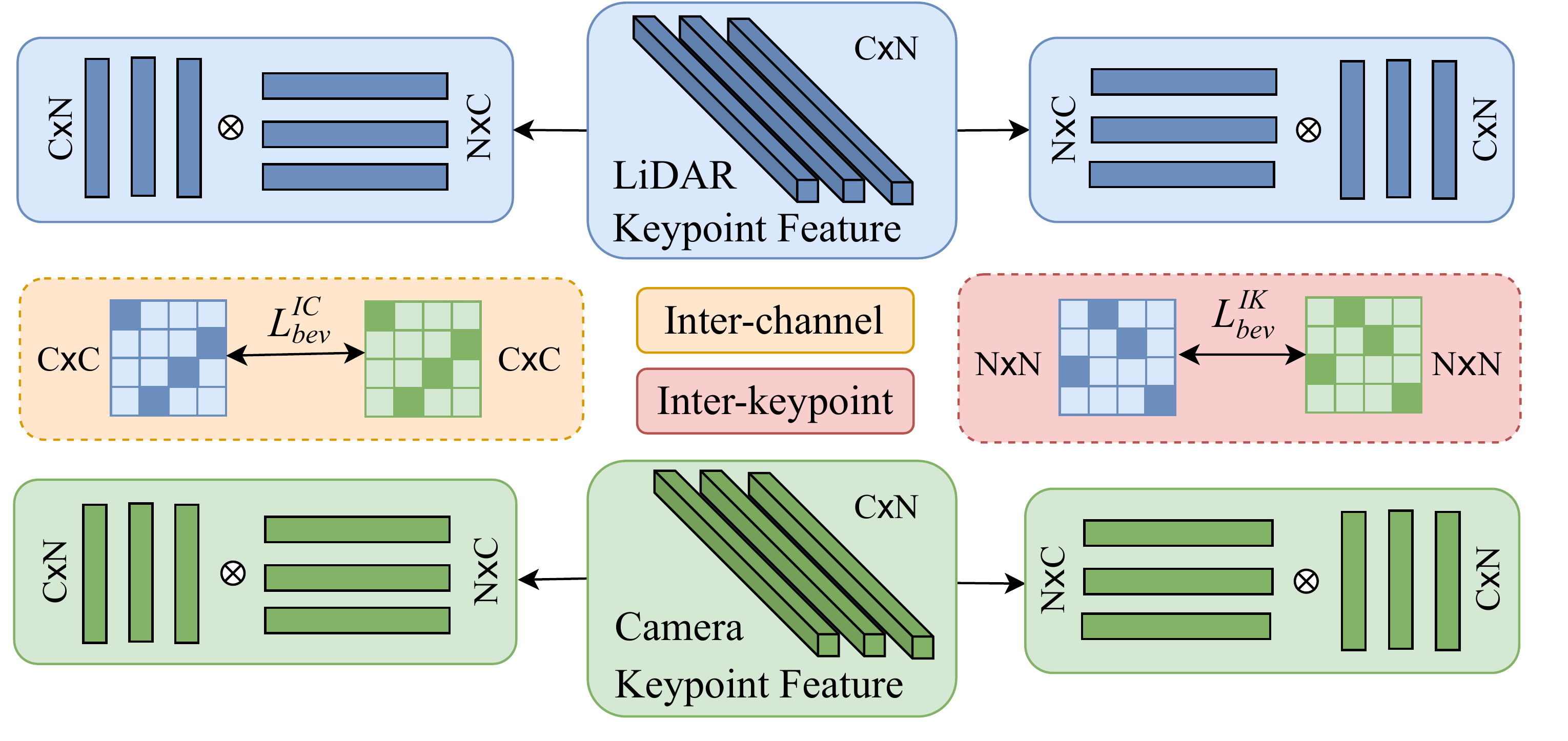}
    \caption{\textit{Details of Innter-feature BEV Distillation.} For each foreground area in BEV space, we represent each target feature by a set of key points and conduct feature distillation in both inter-channel and inter-keypoint manners.
    }
   \label{fig:structure_attn}
\end{figure}

\subsection{\textbf{Inner-feature BEV Distillation}} 
\label{sec:Inner-feature BEV Distillation}

Beyond depth supervision for low-level spatial cues, our TiGDistill-BEV employs inner-geometry learning to extract high-level BEV semantics from pre-trained teacher detectors. Previous works for BEV distillation directly force the student to imitate the teacher's features point-to-point in the BEV space which can be problematic due to potential noise in background areas (arising from sparse point clouds). Although BEVDistill utilizes foreground masks to alleviate this issue, such dense feature distillation still cannot provide focused and effective guidance to the student model. On the other hand, forcing the BEV features to be completely consistent between two modalities is sub-optimal considering the semantic gap. To overcome these limitations, we propose an inner-feature BEV distillation (Figure. \ref{fig:structure_attn}) consisting of inter-channel and inter-keypoint learning schemes, which conducts attentive target features distillation and relieve the cross-modal semantic gap.

\subsubsection{\textbf{Target Keypoint Extraction}}
To distill knowledge from teacher detectors within sparse foreground regions, we focus on extracting the BEV area for each target object and represent it using a series of keypoint features.  Given the ground truth 3D bounding box for each target, we slightly enlarge it in the BEV space to cover the entire foreground area, \eg, object contours and edges. We then uniformly sample $N$ keypoints within BEV bounding box and use bilinear interpolation to obtain the keypoint features from the encoded BEV representations. From both camera-based $F^{cam}_{\rm bev}$ and  $F^{teacher}_{\rm bev}$, we respectively extract the keypoint features for all $M$ object targets as $\{f_j^{cam}, f_j^{teacher}\}_{j=1}^M$, where $f_j^{cam}, f_j^{teacher} \in {\mathbb{R}^{N\times C}}$. These BEV keypoints effectively capture the part-wise features and the inner-geometry semantics of foreground targets.

\subsubsection{\textbf{Inter-channel BEV Distillation}}
Taking the $j$-th object target as an example, we first apply an inter-channel BEV distillation, which guides the student keypoint features to mimic the channel-wise relationships of the teacher. Such inter-channel signals imply the overall geometric semantics of each object target. Compared with the previous channel-by-channel supervision, our inter-channel distillation can preserve the distinctive aspects of the two modalities, while effectively transferring the well pre-trained knowledge of teacher detectors. Specifically, we calculate the inter-channel similarities of both camera-based and LiDAR-based keypoint features, formulated as follows:
\begin{equation}
    A_j^{cam} = f_j^{cam} {f_j^{cam}}^{\top}; \  A_j^{teacher} = f_j^{teacher} {f_j^{teacher}}^{\top}
\end{equation}
where $A_j^{cam}$, $A_j^{teacher} \in {\mathbb{R}^{C\times C}}$ denote the feature relationships between different $C$ channels of the diverse modalities. For all $M$ objects in a scene, we adopt L2 loss between the two inter-channel similarities for feature distillation as:
\begin{equation}
    \mathcal{L}_{\rm{bev}}^{{IC}}= \sum_{j=1}^{M} ||A_j^{teacher}-A_j^{cam}||_2.
    \label{eq:fm}
\end{equation}

\subsubsection{\textbf{Inter-keypoint BEV Distillation}}
\label{sec:anchor}
While inter-channel distillation guides the student model to learn the channel-wise diversity from the teachers, it neglects the inner correlations between different keypoints within a target, which are crucial for understanding the local geometries among different foreground parts, \eg, the front and rear of cars. To this end, we leverage the inter-keypoint correlations of teachers' BEV features and transfer such inner-geometry semantics into camera-based detectors. Analogous to the aforementioned inter-channel module, for the $j$-th target object, we calculate the inter-keypoint similarities in a transposed manner for both modalities as follows:
\begin{equation}
    B_j^{cam} = {f_j^{cam}}^{\top} {f_j^{cam}};\ B_j^{teacher} = {f_j^{teacher}}^{\top} {f_j^{teacher}}
\end{equation}
where $B_j^{cam}, B_j^{teacher} \in {\mathbb{R}^{N\times N}}$ denote the feature relationships between different $N$ keypoints respectively for camera and teachers'. We also adopt L2 loss for all $M$ targets as:
\begin{equation}
    \mathcal{L}_{\rm{bev}}^{{IK}}= \sum_{j=1}^{M} ||B_j^{teacher}-B_j^{cam}||_2.
    \label{eq:fm}
\end{equation}
Subsequently, the distillation loss for inter-channel and inter-keypoint features is formulated as:
\begin{equation}
    \mathcal{L}_{\rm{bev}}^{{D}}=
    \mathcal{L}_{\rm{bev}}^{{IC}}+
    \mathcal{L}_{\rm{bev}}^{{IK}},
    \label{eq:seg}
\end{equation}
where the two terms are orthogonal for the channel-wise feature diversity and keypoint-wise semantic correlations, respectively.

\subsection{\textbf{Overall Loss}}
\label{sec:overall_loss}
To summarize, we enhance the student camera-based detector by leveraging target inner-geometry from two complementary aspects, an inner-depth supervision for low-level signals and an inner-feature BEV distillation for high-level semantics which contribute two losses as $\mathcal{L}^R_{\rm{depth}}$ and $\mathcal{L}_{\rm{bev}}$. Combined with the original losses of depth supervision from BEVDepth $\mathcal{L}^A_{\rm{depth}}$, and 3D detection $\mathcal{L}_{\rm{det}}$, the overall loss of our TiGDistill-BEV is formulated as follows:
\begin{equation}
    \mathcal{L}_{\rm{TiG}}=
    \mathcal{L}_{\rm{det}}+
    \mathcal{L}^A_{\rm{depth}}+
    \mathcal{L}^R_{\rm{depth}}+
    \mathcal{L}^{IC}_{\rm{bev}}+
    \mathcal{L}^{IK}_{\rm{bev}}
    \label{eq:seg}
\end{equation}

\section{EXPERIMENTAL RESULTS}
\label{sec:experiment}
In this section, we present the details of the dataset and introduce our implementation settings, conducting a series of experiments with ablation studies to show the effectiveness of our framework TiGDistill-BEV.

\subsection{Dataset and Evaluation Metrics}
\noindent\textbf{nuScenes dataset.} {We evaluate our TiGDistill-BEV on nuScenes\cite{b6} dataset, which is one of the most popular large-scale outdoor public datasets for autonomous driving. It provides synced data captured from a 32-beam LiDAR at 20Hz and six cameras covering 360-degree horizontally at 12Hz. We adopt the official evaluation toolbox, which reports the nuScenes Detection Score (NDS) and mean Average Precision (mAP), along with mean Average Translation Error (mATE), mean Average Scale Error (mASE), mean Average Orientation Error (mAOE), mean Average Velocity Error (mAVE), and mean Average Attribute Error (mAAE). It consists of 700, 150, 150 scenes for training, validation and testing, respectively.}

\noindent\textbf{Evaluation Metrics (Detection).} The evaluation metric for nuScenes is totally different from KITTI and they propose to use mean Average Precision (mAP) and nuScenes detection score (NDS) as the main metrics. Different from the original mAP defined in \cite{everingham2010pascal}, nuScenes consider the BEV center distance with thresholds of \{0.5, 1, 2, 4\} meters, instead of the IoUs of bounding boxes. NDS is a weighted sum of mAP and other metric scores, such as average translation error (ATE) and average scale error (ASE). For more details about the evaluation metric please refer to \cite{b6}.

\begin{table*}[ht]
\centering
\caption{\textit{Performance on the nuScenes Val Set.} 'C' and 'L' denote the camera-based and LiDAR-based methods, respectively. * denotes our implementation using BEVDet~\cite{b19} codebase, and $^\textit{T}$ indicate the teacher model employed.
}
{
\begin{tabular}{l|c|c|c|ll|lllll}

\toprule
Method          & Modality & Backbone & Resolution & mAP↑  &NDS↑& mATE↓ & mASE↓ & mAOE↓ & mAVE↓ & mAAE↓   \\ \midrule
\textbf{CenterPoint$^\textit{T}$}& \textbf{L} &VoxelNet   & -   & 56.4 & 64.6& 29.9 & 25.4 & 33.0 & 28.6 & 19.1  \\ \midrule
MonoDETR             & C  &ResNet-101 & 900 $\times$ 1600 & 37.2& 43.4 & 67.6 & 25.8 & 42.9 & 125.3 & 17.6  \\
DETR3D          & C &ResNet-101  & 900 $\times$ 1600 & 30.3& 37.4 & 86.0 & 27.8 & 43.7 & 96.7 & 23.5  \\
PETR            & C  &ResNet-101    & 512 $\times$ 1408 & 35.7 & 42.1& 71.0 & 27.0 & 49.0 & 88.5 & 22.4  \\
BEVFormer       & C  &ResNet-101   & 900 $\times$ 1600 & 41.6 & 51.7& 67.3 & 27.4 & 37.2 & 39.4 & 19.8   \\
PETRv2            & C  &ResNet-101    & 640 $\times$ 1600 & 42.1& 52.4 & 68.1 & 26.7 & 35.7 & 37.7 & 18.6  \\
Sparse4D            & C  &ResNet-101    & 900 $\times$ 1600 & 43.6& 54.1 & 63.3 & 27.9 & 33.3 & 31.7 & 17.7  \\
\midrule

BEVDet & C &Swin-T   & 256 $\times$ 704 & 31.2& 39.2 & 69.1 & 27.2 & 52.3 & 90.9 & 24.7  \\ 
\rowcolor{gray!12} \textbf{+ TiGDistill-BEV}     &L $\rightarrow$ C  &Swin-T  & 256 $\times$ 704 & \textbf{33.4}(\textbf{\textcolor{blue}{+2.2}}) & \textbf{42.1}(\textbf{\textcolor{blue}{+2.9}})& 64.7(\textcolor{blue}{-4.4}) & 27.0(\textcolor{blue}{-0.2}) & 55.4(\textcolor{blue}{+3.1}) & 77.7(\textcolor{blue}{-13.2}) & 21.1(\textcolor{blue}{-3.6})  \\ 

\midrule
BEVDet4D & C &ResNet-101   & 512 $\times$ 1408 & 37.0 & 49.2& 66.7& 27.2& 47.5 &33.1 &18.2  \\
\rowcolor{gray!12} \textbf{+ TiGDistill-BEV}   & L $\rightarrow$ C &ResNet-101  & 512 $\times$ 1408 & \textbf{41.2}(\textbf{\textcolor{blue}{+4.2}})& \textbf{52.0}(\textbf{\textcolor{blue}{+2.8}}) & 60.8(\textcolor{blue}{-5.9})& 27.1(\textcolor{blue}{-0.1}) & 45.1(\textcolor{blue}{-2.4}) & 34.1(\textcolor{blue}{+1.0}) & 18.7(\textcolor{blue}{+0.5})  \\ 

\midrule
BEVDepth$^*$ & C &ResNet-101   & 512 $\times$ 1408 & 41.6 & 52.1&60.5 & 26.8 & 45.5 & 33.3 & 20.3  \\
\rowcolor{gray!12}   \textbf{+ TiGDistill-BEV}   & L $\rightarrow$ C & ResNet-101  & 512 $\times$ 1408  & \textbf{44.0}(\textbf{\textcolor{blue}{+2.4}})& \textbf{54.4}(\textbf{\textcolor{blue}{+2.3}}) & 57.0(\textcolor{blue}{-3.5}) & 26.7(\textcolor{blue}{-0.1}) & 39.2(\textcolor{blue}{-6.3}) & 33.1(\textcolor{blue}{-0.2}) & 20.1(\textcolor{blue}{-0.2}) \\ 
\bottomrule
\end{tabular}
}

\label{tab:nus_val_sota}
\end{table*}

\begin{table*}[t!]
\centering
\caption{\textit{Performance on the nuScenes Test Set.} 'C' denotes the camera-based method, which refers to the input during inference. * denotes our implementation are same as BEVDistill, and $\dagger$ is reported by DistillBEV. 
}
{
\begin{tabular}{l|c|c|c|ll|lllll}

\toprule
Method          & Modality & Backbone & Resolution&  mAP↑  &NDS↑& mATE↓ & mASE↓ & mAOE↓ & mAVE↓ & mAAE↓   \\ \midrule
\textbf{BEVFusion$^\textit{T}$ }& \textbf{L + C} & -  & -          & 64.66 & 69.20&28.64&25.78&32.55&25.07&19.31\\\midrule

DETR3D          & C &VoVNet-99&900 $\times$ 1600  & 41.2& 47.9 & 64.1 & 25.5 & 39.4 & 84.5 & 13.3  \\
PETR            & C&VoVNet-99&900 $\times$ 1600   & 44.1 & 50.4& 59.3 & 24.9 & 38.4 & 80.8 & 13.2  \\
BEVFormer       & C &VoVNet-99&900 $\times$ 1600  & 48.1 & 56.9& 58.2 & 25.6 & 37.5 & 37.8 & 12.6   \\
PETRv2            & C &VoVNet-99&640 $\times$ 1600  & 49.0& 58.2 & 56.1 & 24.3 & 36.1 & 34.3 & 12.0  \\
BEVDet & C&Swin-B&640 $\times$ 1600   & 42.4 & 48.8& 52.4 & 24.2 & 37.3 & 95.0 & 14.8  \\
BEVDet4D & C &Swin-B&640 $\times$ 1600  & 45.1 & 56.9& 51.1 & 24.1 & 38.6 & 30.1 & 12.1  \\
Unidistill&L $\rightarrow$ C&ResNet-50& 256 $\times$ 704  & 28.9 & 38.4& 65.9 & 25.9 & 51.4 & 106.4 & 17.0 \\
X$^3$KD&L $\rightarrow$ C&ResNet-101& 640 $\times$ 1600  & 45.6 & 56.1& 50.6 & 25.3 & 41.4 & 36.6 & 13.1 \\
\midrule
BEVDepth$\dagger$ & C&Swin-B&640 $\times$ 1600  & 48.9 & 59.0& - & - & - & - & - \\

+ DistillBEV & L + C $\rightarrow$ C&Swin-B&640 $\times$ 1600  & 52.5 & 61.2& - & - & - & - & - \\
\midrule
BEVDepth$^*$ & C&ConvNeXt-B&640 $\times$ 1600  & 49.1 & 58.9& 48.4 & 24.5 & 37.7 & 32.0 & 13.2  \\

+ BEVDistill & L $\rightarrow$ C&ConvNeXt-B&640 $\times$ 1600  & 49.8 & 59.4& 47.2 & 24.7 & 37.8 & 32.6 & 12.5  \\
\rowcolor{gray!12}   \textbf{+ TiGDistill-BEV}   & L $\rightarrow$ C   &ConvNeXt-B&640 $\times$ 1600& \textbf{53.2}(\textbf{\textcolor{blue}{+4.1}})& \textbf{61.9}(\textbf{\textcolor{blue}{+3.0}}) & 45.0(\textcolor{blue}{-3.4}) & 24.4(\textcolor{blue}{-0.1}) & 34.3(\textcolor{blue}{-3.4}) & 30.6(\textcolor{blue}{-1.4}) & 13.2(\textcolor{blue}{-0.0}) \\ 
\rowcolor{gray!12}   \textbf{+ TiGDistill-BEV}   & L + C$\rightarrow$ C   &ConvNeXt-B&640 $\times$ 1600& \textbf{53.9}(\textbf{\textcolor{blue}{+4.8}})& \textbf{62.8}(\textbf{\textcolor{blue}{+3.9}}) & 43.9(\textcolor{blue}{-4.5})& 24.1(\textcolor{blue}{-0.4}) & 32.8(\textcolor{blue}{-4.9}) & 30.0(\textcolor{blue}{-2.0}) & 12.8(\textcolor{blue}{-0.4})  \\
\bottomrule
\end{tabular}
}

\label{tab:nus_test_sota}
\end{table*}

\noindent\textbf{Evaluation Metrics (Depth Estimation).} 
To assess and compare the performance of different depth estimation networks, a widely accepted evaluation methodology, proposed in \cite{eigen2014depth} with five evaluation indicators: \textbf{ RMSE, RMSE log, Abs Rel, Sq Rel, Accuracies}. These indicators are formulated as:

\begin{itemize}
	\item $\textbf{RMSE} = \sqrt{\frac{1}{|N|}\sum_{i\in N}\parallel d_{i}-d_{i}^{*} \parallel^{2}}$,
	
	\item $\textbf{RMSE log} = \sqrt{\frac{1}{|N|}\sum_{i\in N}\parallel \log (d_{i})- \log (d_{i}^{*}) \parallel^{2}}$,
	
	\item $\textbf{Abs Rel} = \frac{1}{|N|}\sum_{i\in N}\frac{\mid d_{i}-d_{i}^{*}\mid}{d_{i}^{*}}$,\quad
	\item $\textbf{Sq Rel} = \frac{1}{|N|}\sum_{i\in N}\frac{\parallel d_{i}-d_{i}^{*}\parallel^{2}}{d_{i}^{*}}$,
	
	\item \textrm{\textbf{Accuracies:} $\%$ of $d_{i}$ s.t. } $\max(\frac{d_{i}}{d_{i}^{*}}, \frac{d_{i}^{*}}{d_{i}}) = \delta < thr$,
\end{itemize}
where $d_{i}$ is the predicted depth value of pixel $i$, and $d_{i}^{*}$ stands for the ground truth of depth. Besides, $N$ denotes the total number of pixels with real-depth values, and $thr$ denotes the threshold.

\subsection{Implementation Details.}
We implement our TiGDistill-BEV using the BEVDet code base on 8 NVIDIA A100 GPUs.
The pre-trained teacher model CenterPoint, BEVFusion with voxel sizes $[0.1,0.1,0.2]$ and $[0.1,0.1,8]$ for PillarNext. The camera-based students include BEVDepth, BEVDet and BEVDet4D.
Referring to BEVDepth, we additionally add the dense depth supervision on top of BEVDet and BEVDet4D besides our TiGDistill-BEV. We follow their official training settings as default, including data augmentation (random flip, scale and rotation), training schedule (2x), and others (AdamW optimizer, 2e-4 learning rate and batch size 8). The CBGS strategy~\cite{b56} is employed in Table. 1. We implement TiGDistill-BEV on BEVDepth with ConvNeXt-base~\cite{b63} backbone and input images of $640 \times 1600$ in nuScenes test set. During distillation, we utilize the freeze the teacher model and train the student model for 20 epochs with CBGS and evaluate without test time augmentation. For other results, we do not utilize CBGS to better highlight the significance.

\subsection{Main Results}
\label{sec:main_result}
The proposed framework has been evaluated on nuScenes benchmark for both “val” and “test” splits. 
\subsubsection{On nuScenes Val Set} To demonstrate the effectiveness of the proposed TiGDistill-BEV framework, we conducted experiments on the nuScenes validation dataset.  We employed BEVDet, BEVDet4D, and BEVDepth as baseline detectors. Table. \ref{tab:nus_val_sota} presents the results, showing significant improvements in both mean Average Precision (mAP) and NuScenes Detection Score (NDS) across all baselines. Specifically, TiGDistill-BEV achieved gains of 2.2\%, 4.2\%, and 2.4\% mAP and 2.9\%, 2.8\%, and 2.3\% NDS, respectively.  Furthermore, TiGDistill-BEV consistently produced lower error metrics (e.g., mATE) compared to the baseline detectors, indicating its superiority in object localization. Notably, our framework achieved a 13.2\% reduction in mean Average Velocity Error (mAVE) compared to BEVDepth. This result demonstrates the successful distillation of prior knowledge from LiDAR data to the camera-based framework. The Centerpoint\cite{yin2021center} served as the LiDAR-based detector teacher network for these validation experiments.
\subsubsection{On nuScenes Test Set}We further evaluated TiGDistill-BEV on the nuScenes test set, utilizing \textbf{BEVFusion}\cite{liu2023bevfusion} as the teacher model to provide diverse multi-modal supervision. As shown in Table.\ref{tab:nus_test_sota}, TiGDistill-BEV achieves state-of-the-art performance, surpassing other camera-based 3D detectors with \textbf{62.8\% }NDS and \textbf{53.9\%} mAP. The overall performance gain of 3.9\% NDS and 4.8\% mAP compared to the baseline BEVDepth.  Using a LiDAR-based teacher model yielded improvements of 4.1\% mAP and 3.0\% NDS.  Furthermore, incorporating both LiDAR and camera modalities as the teacher model resulted in additional gains of 0.7\% and 0.9\% NDS, respectively.  Notably, our framework outperforms BEVDistill by 3.4\% NDS and 4.1\% mAP. These results underscore the effectiveness of our target inner-geometry learning scheme in enhancing multi-view 3D object detection.

\subsection{Ablation Study}
\label{sec:ablation}
In this section, we carried out extensive experiments to analyze and understand the proposed components and related design choices in TiGDistill-BEV. Unless otherwise specified, the following experiments employed Centerpoint and BEVDepth as the teacher and student model, respectively.

\begin{table}[h]
\centering
\caption{\textit{Ablation Study of Target Inner-geometry Learning. }$\mathcal{L}^R_{\rm{depth}}$ and $\mathcal{L}_{\rm{bev}}$  denote the losses of inner-depth supervision and inner-feature BEV distillation, respectively.}

    {
    \begin{tabular}{cc|cc}
    \toprule
          $\mathcal{L}^R_{\rm{depth}}$&$\mathcal{L}_{\rm{bev}}$& mAP&NDS  \\
          \midrule
                        &    & 32.9         & 43.1 \\ 
             \checkmark &    & 33.9(\textcolor{blue}{+1.0})         & 44.0(\textcolor{blue}{+0.9}) \\
                        &\checkmark & 35.9(\textcolor{blue}{+3.0})         & 45.4(\textcolor{blue}{+2.3}) \\
              \checkmark&\checkmark & \textbf{36.6}(\textcolor{blue}{+3.7})         & \textbf{46.1}(\textcolor{blue}{+3.0}) \\
    \bottomrule
    \end{tabular}}

    \label{tab:main_ablation}
\end{table}
\begin{table}[h]
   \centering
    \caption{\textit{Ablation Study of Inner-depth Supervision.} We compare different settings for relative depth value calculation and depth reference selection.}   
    {
    \begin{tabular}{c|c|cc}
    \toprule
          Setting &Depth Reference & mAP&NDS  \\
          \midrule
          BEVDepth & - & 32.9 & 43.1 \\
          \midrule
        \multirow{2}{*}{All-to-Certain}& 3D Center  & 35.8         & 45.2 \\ 
             &2D Center  & 35.8        & 45.2 \\
             \midrule
         \multirow{2}{*}{All-to-Adaptive}&Highest Conf & 35.7         & 45.5 \\
              
              &Smallest Error & \textbf{36.6}         & \textbf{46.1} \\
    \bottomrule
    \end{tabular}}

    \label{tab:relative_ablation}
\end{table}

\subsubsection{Inner-geometry Learning} 
The individual effectiveness of the two main components can be examined by only equipping one of them.
Table. \ref{tab:main_ablation} examines the individual contributions of inner-depth supervision and inner-feature BEV distillation. Introducing inner-depth supervision alone yields gains of +1.0\% mAP and +0.9\% NDS. Similarly, employing inner-feature BEV distillation independently results in gains of +3.0\% mAP and +2.3\% NDS. Combining both components further boosts performance to +3.7\% mAP and +3.0\% NDS, demonstrating the two proposed objectives can collaborate for better performance, highlighting their complementary nature.

\subsubsection{Inner-depth Supervision}
In order to calculate the relative depth values within foreground targets, we compare two paradigms concerning the relationships among different inner points, 1) \textbf{\emph{All-to-Certain}} calculates the relative depth from all sampled points to a certain reference point, such as the projected center of 3D bounding box or the center of 2D bounding box. 2) \textbf{\emph{All-to-Adaptive}} dynamically selects the reference pixel with the highest confidence across all depth bins or the smallest depth error to the ground truth (Ours). As shown in Table. \ref{tab:relative_ablation}, \textbf{\emph{All-to-Adaptive with smallest depth errors}} obtains the best improvement, which indicates the dynamic depth reference point can flexibly adapt to different targets for inner-geometry learning. 

\begin{table}[h]
\centering
\caption{\textit{Ablation Study of Inner-feature BEV Distillation.} $\mathcal{L}^{IC}_{\rm{bev}}$ and $\mathcal{L}^{IK}_{\rm{bev}}$ denote the losses of inter-channel and inter-keypoint distillation, respectively.}
    {
    \begin{tabular}{c|cc|cc}
    \toprule
    $\mathcal{L}^R_{\rm{depth}}$ & $\mathcal{L}^{IC}_{\rm{bev}}$&$\mathcal{L}^{IK}_{\rm{bev}}$& mAP$\uparrow$&NDS$\uparrow$  \\
          \midrule
         \multirow{4}{*}{\checkmark}        &       &           & 32.9         & 43.1 \\ 
            & \checkmark &           & 34.2(\textcolor{blue}{+1.3})         & 44.4(\textcolor{blue}{+1.3}) \\
            &            &\checkmark & 35.8(\textcolor{blue}{+2.9})         & 45.2(\textcolor{blue}{+2.1}) \\
          &    \checkmark&\checkmark & \textbf{36.6}(\textcolor{blue}{+3.7})         & \textbf{46.1}(\textcolor{blue}{+3.0}) \\
    \bottomrule
    \end{tabular}}

    \label{tab:saf_ablation}
\end{table}
\begin{table}[h]
\centering
    \caption{Target Depth Prediction Comparison.}
    \vspace{-3.5mm}
    (\textit{The smaller the value, the better the result.)}
    {
    \begin{tabular}{c|cc}
    \toprule

        Metric & BEVDepth$^*$ & + TiGDistill-BEV \\ \midrule
        SILog$\downarrow$ & 29.306 & \textbf{15.897} (\textcolor{blue}{$\uparrow$45\%})\\
        Abs Rel$\downarrow$ & 0.217&\textbf{0.091} (\textcolor{blue}{$\uparrow$56\%})\\
        Sq Rel$\downarrow$ & 1.103&\textbf{0.232} (\textcolor{blue}{$\uparrow$79\%})\\
        log10$\downarrow$ & 0.079&\textbf{0.038} (\textcolor{blue}{$\uparrow$52\%})\\
        RMSE$\downarrow$ & 3.167&\textbf{2.077} (\textcolor{blue}{$\uparrow$35\%})\\

    \bottomrule
    \end{tabular}
    }
    \label{tab:depth_error}
\end{table}

\begin{figure}[h]
    \centering
    \includegraphics[width=\columnwidth]{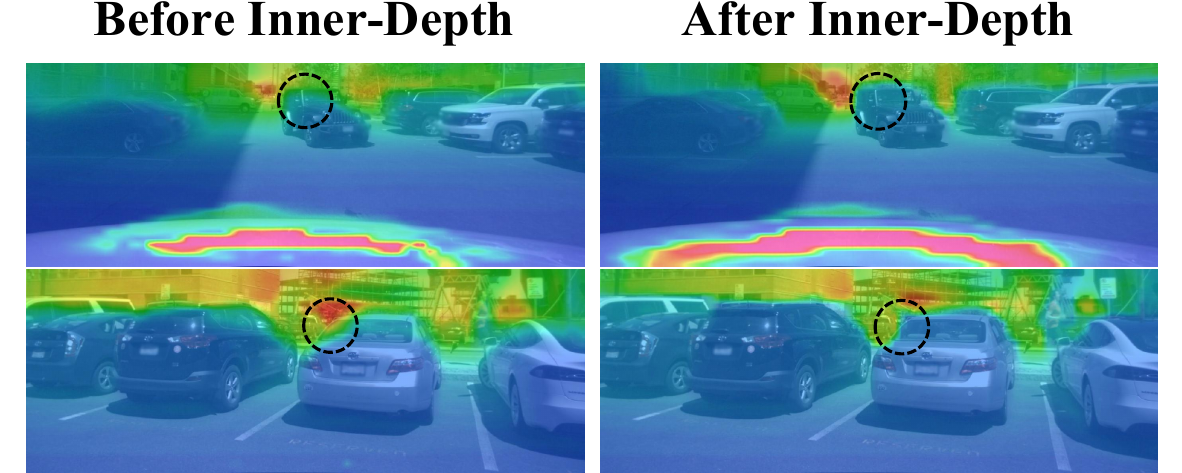}
    \caption{\textit{Visualization of Predicted Depth Maps.} We show the predicted depth maps before and after the inner-depth supervision, respectively, to better understand the impacts of this module.}
    \label{fig:ref_compare}
\end{figure}

\subsubsection{Inner-feature BEV Distillation}
Our TiGDistill-BEV explores the BEV feature distillation from two perspectives, \textit{inter-channel} and \textit{inter-keypoint}. As shown in Table. \ref{tab:saf_ablation}, both distillation sub-modules contribute positively to the final performance, with NDS improvements of +1.3\% and +2.1\%,  respectively. This well illustrates the importance of learning inner-geometry semantics within different foreground targets in BEV space. Further combining them can benefit the performance by \textbf{+3.7\%} and \textbf{+3.0\%} of mAP and NDS. A more comprehensive comparison of distillation methods is shown in Table. \ref{tab:cccc} of \cref{comapare_distill}.

\subsubsection{Target Depth Estimation}
In the Table. \ref{tab:depth_error}, we compare the performance of our proposed TiGDistill-BEV against the baseline BEVDepth model for depth estimation. Specifically, it achieves a 45\%, 56\%, 79\%, 52\% and a 35\% improvement at the indicator of SILog, Abs Rel, Sq Rel, log10 and RMSE, respectively. Which is consistent with the visualization of predicted depth maps in Figure. \ref{fig:ref_compare} and facilitates target positioning in BEV space.  The results clearly indicate that our method effectively enhances the accuracy of depth estimation compared to the baseline model, achieving lower error values across all metrics.

\begin{table}[h]
\centering
    \caption{\textit{Performance with different teacher models.}  * denote teacher model implementation of ours on the student model. Which takes images with a size of 256 × 704 as input.}
\setlength{\tabcolsep}{1mm}
{
\begin{tabular}{l|l|l|ll}
\toprule
          Mode &Method& Modality & mAP$\uparrow$&NDS$\uparrow$  \\
          \midrule
           student &BEVDepth$\dagger$& C & 35.7  & 48.1     \\
            \midrule
          \multirow{7}{*}{teacher}&Centerpoint& L & 56.4  & 64.6 \\ 
         &Centerpoint$^*$& L $\rightarrow$  C& \textbf{38.3} (\textcolor{blue}{+2.6})  & \textbf{49.8} (\textcolor{blue}{+1.7}) \\ 
          \cmidrule(lr){2-5}
           &PillarNext & L  & 60.6 &67.4 \\
           &PillarNext$^*$ & L $\rightarrow$  C & \textbf{38.7} (\textcolor{blue}{+3.0}) &\textbf{50.4} (\textcolor{blue}{+2.1}) \\
           \cmidrule(lr){2-5}
          &BEVFusion & L + C& {64.6} &{69.2}  \\
        &BEVFusion$^*$ & L + C$\rightarrow$  C& {\textbf{39.6} (\textcolor{blue}{+3.9})} &\textbf{51.1} (\textcolor{blue}{+3.0})  \\
    \bottomrule
    \end{tabular}}

    \label{tab:teacher}
\end{table}

\begin{table}[h]
\centering
    \caption{\textit{Comparison with BEVDistill.} $\dagger$ and * denote the implementation of BEVDistill and ours, respectively. We present the performance improvement of the learning methods correspondingly to their implemented baselines.}
\setlength{\tabcolsep}{1mm}
{
\begin{tabular}{l|ll}
\toprule
          Method& mAP$\uparrow$&NDS$\uparrow$  \\
          \midrule
           BEVDepth$\dagger$ &31.1       &  43.2     \\
            \rowcolor{gray!12}+ BEVDistill& 33.2 (\textcolor{blue}{+2.1\%}) &45.4 (\textcolor{blue}{+2.2\%}) \\
            \midrule
          BEVDepth$^*$& 32.9 & 43.1 \\ 
          
           \rowcolor{gray!12}+ Naive Distill& 33.8 (\textcolor{blue}{+0.9\%}) &43.4 (\textcolor{blue}{+0.3\%}) \\
           
          \rowcolor{gray!12}\textbf{+ Inner-feature Distill}& \textbf{35.9 (\textcolor{blue}{+3.0\%})} &\textbf{45.4 (\textcolor{blue}{+2.3\%})} \\
          \rowcolor{gray!12}\textbf{+ TiGDistill-BEV}& \textbf{36.6 (\textcolor{blue}{+3.7\%})} &\textbf{46.1 (\textcolor{blue}{+3.0\%})} \\
    \bottomrule
    \end{tabular}}

    \label{tab:cccc}
\end{table}
\subsubsection{Performance of difference teacher models} We further validate the experiments by incorporating multiple different teacher models as shown in Table. \ref{tab:teacher}. Taking BEVDepth as camera-based student model and leveraging LiDAR data through Centerpoint and PillarNext as teacher models leads consistently gain 2.6\% mAP and 1.7\% NDS, while PillarNext* gains 3.0\% mAP and 2.1\% NDS, respectively. Moreover, BEVFusion, which utilizes both LiDAR and camera data for the distillation student model yields the best results: 39.6\% mAP (a gain of 3.9\%) and 51.1\% NDS (a gain of 3.0\%). This underscores the effectiveness and robustness of transferring knowledge from teacher models trained on richer sensor modalities to enhance the performance of a student model that relies solely on camera input.

\begin{table*}[t]
\centering
    \caption{\textit{Ablation Study of Image Backbones and Temporal Information.} CenterPoint and BEVDepth~\cite{b7} are adopted as the teacher and student models.}
{
\begin{tabular}{c|c|c|c|cc}
\toprule
          Backbone&Resolution&Multi-frame&Method& mAP&NDS  \\
          \midrule
          VoxelNet & - & \checkmark&Teacher & 56.4 & 64.6 \\\midrule
          \multirow{4}{*}{ResNet-18}&\multirow{4}{*}{$256\times 704$}&& Student& 26.0  & 29.5 \\
               & & &    + TiGDistill-BEV& \textbf{29.4 (\textcolor{blue}{+3.4\%})}  & \textbf{33.5 (\textcolor{blue}{+4.0\%})}\\
              \cmidrule(lr){3-6}
                & & \multirow{2}{*}{\checkmark}&Student & 28.5 & 40.5 \\ 
              & & &  + TiGDistill-BEV& \textbf{32.3 (\textcolor{blue}{+3.8\%})}         & \textbf{43.0 (\textcolor{blue}{+2.5\%})} \\
          \midrule
         \multirow{4}{*}{ResNet-50}&\multirow{4}{*}{$256\times 704$}&&  Student& 29.8  & 32.8 \\
               & & &  + TiGDistill-BEV& \textbf{33.8 (\textcolor{blue}{+4.0\%})}  & \textbf{37.5 (\textcolor{blue}{+4.7\%})}\\
              \cmidrule(lr){3-6}
              & & \multirow{2}{*}{\checkmark}&Student & 32.9 & 43.1 \\ 
              & & &  + TiGDistill-BEV& \textbf{36.6 (\textcolor{blue}{+3.7\%})}         & \textbf{46.1 (\textcolor{blue}{+3.0\%})} \\
          \midrule
          \multirow{4}{*}{ResNet-101}&\multirow{4}{*}{$512\times 1408$}&&  Student& 34.5 & 36.6 \\
               & & &    + TiGDistill-BEV& \textbf{40.3 (\textcolor{blue}{+5.8\%})}        & \textbf{41.6 (\textcolor{blue}{+5.0\%)}} \\
              \cmidrule(lr){3-6}
              & & \multirow{2}{*}{\checkmark}&Student & 39.3 & 48.7 \\ 
              & & &  + TiGDistill-BEV& \textbf{43.0 (\textcolor{blue}{+3.7\%})}         & \textbf{51.4 (\textcolor{blue}{+2.7\%})} \\
               \bottomrule
    \end{tabular}}

    \label{tab:many_ablation}
\end{table*}

\begin{table*}[h]
\centering
\caption{\textit{Comparison with Concurrent Works.} We present the performance improvement of some concurrent works, UniDistill~\cite{b65}, X$^3$KD~\cite{b69}, DistillBEV~\cite{b70}, STXD~\cite{jang2024stxd}, correspondingly to their implemented baselines which uniformly use ResNet-50 as backbone with the image resolution of 256 $\times$ 704.}
{
\begin{tabular}{l|ll|l|ll|l}
\toprule
    Student  & mAP$\uparrow$& NDS$\uparrow$& Method     & mAP$\uparrow$& NDS$\uparrow$& Venue     \\
    \midrule
      \multirow{3}{*}{BEVDet}   & 20.3 & 33.1  & UniDistill & 26.0 & 37.3 & CVPR 2023 \\
        & 30.5 & 37.8  & DistillBEV & 32.7 & 40.7 & ICCV 2023 \\
     & 29.8 & 37.9  & \textbf{TiGDistill-BEV}    & \textbf{33.1} & \textbf{41.1} & Ours      \\
     \midrule
     \multirow{2}{*}{BEVDet4D}  & 32.8 & 45.9  & DistillBEV & 36.3 & 48.4 & ICCV 2023 \\ 
     & 32.2 & 45.1  & \textbf{TiGDistill-BEV}   & \textbf{35.6} & \textbf{47.7} & Ours      \\
     \midrule
     \multirow{4}{*}{BEVDepth} 
     & 35.9 & 47.2  & X$^3$KD$_{modal}$  & 36.8 & 49.4 & CVPR 2023 \\
      & 36.4 & 48.4  & DistillBEV & 38.9 & 49.8 & ICCV 2023 \\
    & 32.9 & 43.1  & STXD (CD)& 37.1 & 48.4 & NeurIPS 2024 \\
      & 35.7 & 48.1  & \textbf{TiGDistill-BEV}    & \textbf{38.3} & \textbf{49.8} & Ours   \\  
    \bottomrule
\end{tabular}}

\label{tab:newwork}
\end{table*}

\begin{table*}[h]
\centering
\caption{\textit{Ablation Study of Small Objects Detection Performance.} We use ResNet-101 as backbone with the image resolution of 512 $\times$ 1408, and train with CBGS. $\mathcal{L}^A_{\rm{depth}}$, $\mathcal{L}^R_{\rm{depth}}$ and $\mathcal{L}_{\rm{bev}}$  denote the losses of depth supervision, inner-depth supervision and inner-feature BEV distillation, respectively.}
{
\begin{tabular}{lll|ccccc}
\toprule
\multirow{2}{*}{$\mathcal{L}^A_{\rm{depth}}$} & \multirow{2}{*}{$\mathcal{L}^R_{\rm{depth}}$}&\multirow{2}{*}{$\mathcal{L}_{\rm{bev}}$}&\multicolumn{5}{c}{mAP$\uparrow$ (\%)  }  \\
 &  &  & pedestrian & motorcycle & bicycle & traffic\_cone & barrier \\
\midrule
        &         &      & 43.3          & 34.5          & 32.4       & 54.8           & 55.2       \\
 \checkmark      &         &      & 44.6          & 35.6          & 32.8       & 56.3           & 55.9       \\
 \checkmark      & \checkmark       &      & 45.4          & 37.3          & 33.7       & 57.8           & 58.8       \\
 \rowcolor{gray!12}\textbf{\checkmark}      & \textbf{\checkmark}       & \textbf{\checkmark}    & \textbf{45.7}(\textcolor{blue}{+2.4})        & \textbf{38.8  }(\textcolor{blue}{+4.3})   & \textbf{37.5}(\textcolor{blue}{+5.1})     & \textbf{58.9}(\textcolor{blue}{+4.1})       & \textbf{57.2}(\textcolor{blue}{+2.0})       \\
    \bottomrule
\end{tabular}}
\label{tab:small_performance}
\end{table*}

\subsubsection{Comparison with BEVDistill}
\label{comapare_distill}
 As shown in Table. \ref{tab:cccc}, we compare our TiGDistill-BEV with BEVDistill under the same setting in Table. \ref{tab:small_performance}. Our approach achieves a higher performance boost for both mAP and NDS. Besides, compare with the naive distillation that directly applies MSE loss to the entire BEV features between camera and LiDAR, where our inner-feature distillation performs better. These well demonstrate the superiority of target inner-geometry learning to foreground-guided dense distillation.

 \begin{table}[h]
\centering
    \caption{\textit{Performance of Different Distance Ranges.} $\mathcal{L}^A_{\rm{depth}}$, $\mathcal{L}^R_{\rm{depth}}$ and $\mathcal{L}_{\rm{bev}}$ denote the depth supervision, inner-depth supervision and inner-feature BEV distillation, respectively}
    
    {
    \begin{tabular}{l|lll|ll}
    \toprule
    Distance (m) &  $\mathcal{L}^A_{\rm{depth}}$ & $\mathcal{L}^R_{\rm{depth}}$ & $\mathcal{L}_{\rm{bev}}$ & mAP$\uparrow$  & NDS$\uparrow$   \\
    \midrule
    {[}0,30)      &         &         &      & 44.2 & 53.7   \\
    {[}0,30)      & \checkmark      &         &      & 46.6 & 55.6   \\
    {[}0,30)      & \checkmark      & \checkmark      &      & 47.4 & 55.8   \\
    \rowcolor{gray!12}\textbf{{[}0,30) }     & \textbf{\checkmark}      & \textbf{\checkmark }     & \textbf{\checkmark  } & \textbf{48.6} & \textbf{56.8 }  \\
    \midrule
    {[}30,60)     &         &         &      & 10.2 & 28.7   \\
    {[}30,60)     & \checkmark      &         &      & 10.3 & 28.5   \\
    {[}30,60)     & \checkmark      & \checkmark      &      & 11.7 & 30.1   \\
    \rowcolor{gray!12}\textbf{{[}30,60)}     & \textbf{\checkmark }     & \textbf{\checkmark }     & \textbf{\checkmark }  & \textbf{12.6} & \textbf{30.2    }    \\
        \bottomrule
    \end{tabular}}

    \label{tab:far_performance}
\end{table}

\subsubsection{Image Backbones and Temporal Information}
We further explore the influence of image backbones and temporal information on our TiGDistill-BEV in Table. \ref{tab:many_ablation}. Our framework brings significant performance improvement consistent over different image backbones. Notably, the target inner-geometry learning schemes can provide positive effect for both single-frame and multi-frame settings. The improvement of mAP ranges from \textbf{+3.4\%} to \textbf{+5.8\%} and the improvement of NDS ranges from \textbf{+2.5\%} to \textbf{+5.0\%}.

\subsubsection{Compare with SOTA distill methods} To account for the different backbone architectures in recent state-of-the-art methods, we evaluated against various recent teacher-student knowledge distillation methods for further validate its effectiveness and robustness as shown in Table\ref{tab:newwork} under same setting. The results demonstrate that our proposed method, while simple, is highly efficient and competitive, achieving performance comparable to state-of-the-art techniques.

\subsubsection{Performance of Small Object Detection}
In the Table. \ref{tab:small_performance}, we explore the impact of each component of our framework on small object detection, which includes pedestrian, motorcycle, bicycle, traffic cone, and barrier classes. 
Each sub-module provides a positive performance boost as shown in the table. And finally, combining all three components yields the best overall performance, with notable gains observed of mAP for pedestrians (+2.4\%), motorcycles (+4.3\%), bicycles (+5.1\%), traffic cones (+4.1\%), and barriers (+2.0\%). The quantitative analysis further confirms the effectiveness of our method in enhancing the detection performance of these small objects with each component.

\begin{figure*}[t]
    \centering
    \includegraphics[width=1\textwidth]{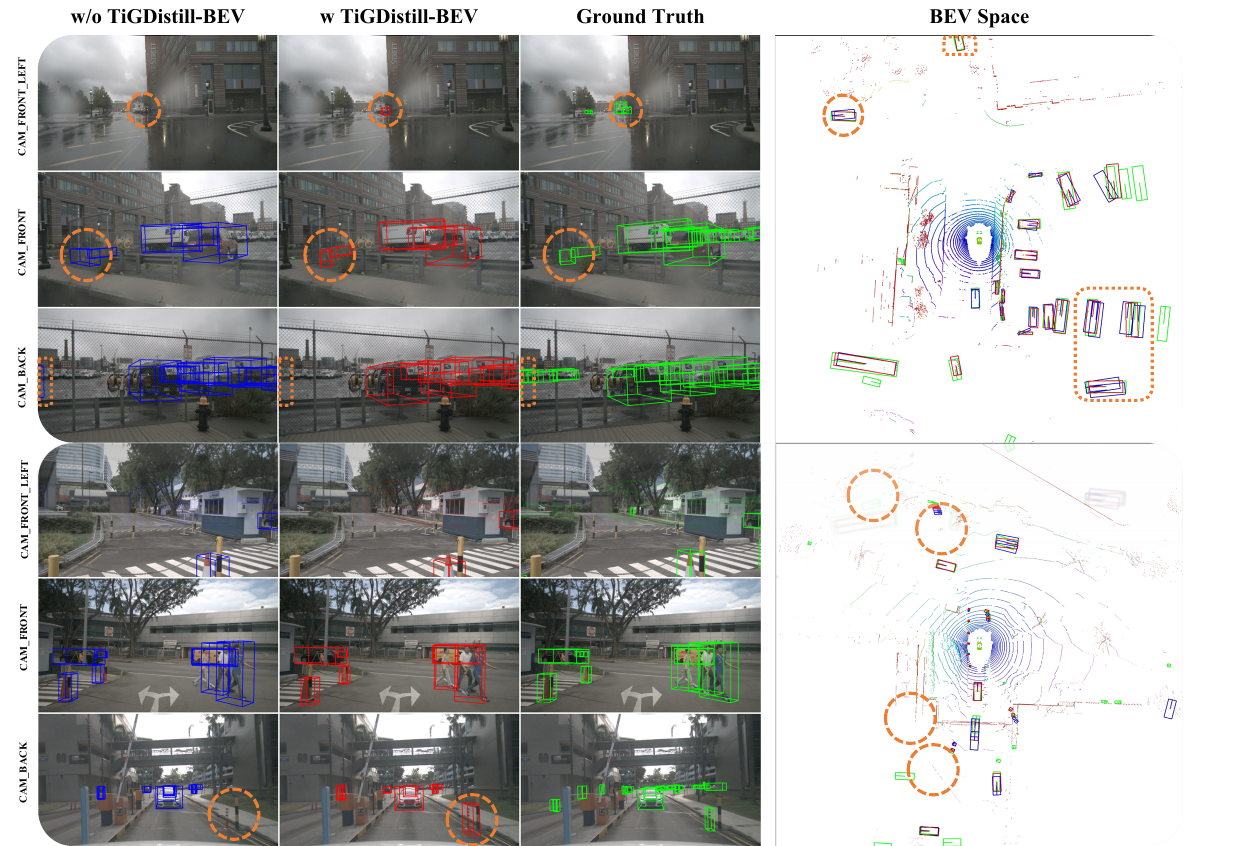}
    \caption{\textbf{Visualization of Detection Results}. From left to right, we show the 3D object detection before and after the TiGDistill-BEV learning schemes, ground-truth annotations, along with the overall BEV-space results.
    }
    \label{fig:app_vis}
\end{figure*}

\subsubsection{Performace of Different Distance Ranges}
The detection of distant objects remains a long-standing challenge due to the sparsity of LiDAR points and the inaccuracy of depth estimation. Here we define 0 to 30 meters as close while 30 to 60 meters as long-distance detection. As seen from the Table. \ref{tab:far_performance}, our method can greatly improve the performance of close detection, and it also works even in the face of long-distance object detection with designed modules. 

\begin{table}[h]
\centering
    \caption{\textit{Performance of Comparison without CBGS~\cite{b56}.} For all methods, we adopt ResNet-101 as the image backbone and $512\times 1408$ as the image resolution. * denotes our implementation.}
{
\begin{tabular}{l|cc}
\toprule
          Method& mAP$\uparrow$&NDS$\uparrow$  \\
          \midrule
          BEVDet$^*$& 27.2 & 29.7 \\
          \rowcolor{gray!12}\textbf{+ TiGDistill-BEV}& \textbf{37.5 (\textcolor{blue}{+10.3\%})} &\textbf{38.8 (\textcolor{blue}{+9.1\%})} \\\midrule
          BEVDet4D$^*$& 33.6 & 43.5 \\
          \rowcolor{gray!12}\textbf{+ TiGDistill-BEV}& \textbf{40.9 (\textcolor{blue}{+7.3\%})} &\textbf{48.9 (\textcolor{blue}{+5.4\%})} \\\midrule

          BEVDepth$^*$& 39.3 & 48.7 \\ 
          \rowcolor{gray!12}\textbf{+ TiGDistill-BEV}& \textbf{43.0 (\textcolor{blue}{+3.7\%})} &\textbf{51.4 (\textcolor{blue}{+2.7\%})} \\
    \bottomrule
    \end{tabular}}

    \label{tab:aaaa}
\end{table}

\subsubsection{Without CBGS~\cite{b56} Strategy} In Table.\ref{tab:aaaa}, we present the results of TiGDistill-BEV without the CBGS training strategy. The performance still improvement of learning target inner-geometry becomes more notable, \textbf{+10.3\%, +7.3\%,} and \textbf{+3.7\%} mAP for the three baselines, which indicates the superior LiDAR-to-camera knowledge transfer of our framework.


\subsection{Visualization.} 
The Figure. \ref{fig:app_vis} visually compares the results of BEVDepth before and after TiGDistill-BEV. The enhanced results showcase (within the orange circles) reduced false positives, improved object localization, and refined 3D bounding box orientations, particularly in challenging areas by our inner-geometry learning.

\subsection{Discussion}
\subsubsection{Improvements of Inner-depth Supervision vs. Inner-feature Distillation} 
Our inner-depth provides fine-grained depth cues for object-level geometry understanding, which is low-level information and implicitly benefits the mAP accuracy.
The inner-feature directly supervises high-level semantics of BEV features, explicitly affecting the recognition performance. They are complementary and can collaborate for better results.

\subsubsection{Limitations and Future Works} 

A notable limitation is the occlusion problem. Due to visual occlusion, the ground truth of the occluded object that we can obtain is limited, which will affect the learning of depth prediction and inner-depth learning. A possible way to alleviate it is to consider multi-frame images for inner-depth supervision of the same object's interior. However, a potential risk here is that if the intrinsic and extrinsic parameters are inaccurate, it will directly affect the coordinate system transformation between multiple frames, thereby affecting the ground truth of depth values.

\section{Conclusion}
In this paper, we propose a novel target inner-geometry learning framework that enables the camera-based detector to inherit the effective foreground geometric semantics from the teacher modality. The inner-depth supervision and inner-feature distillation in BEV space, respectively for learning better low-level structures and high-level semantics to enhance the performance of the student model. 
In future work, our focus will be on exploring a temporal learning strategy for a unified real-world perception system.

\bibliographystyle{IEEEtran}
\bibliography{ref}

\begin{IEEEbiography}[{\includegraphics[width=1in,height=1.25in,clip,keepaspectratio]{{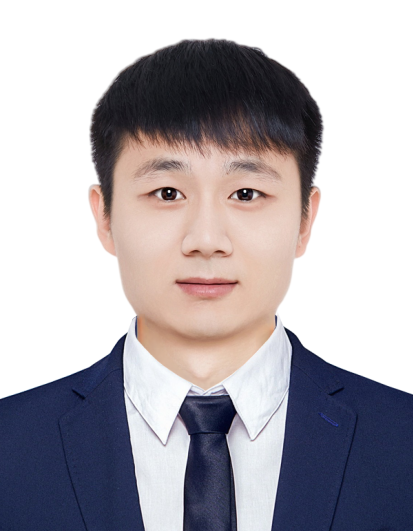}}}]{Shaoqing Xu} received his M.S. degree in transportation engineering from the School of Transportation Science
and Engineering in Beihang University. He is currently working toward the Ph.D. degree in electromechanical engineering with the State Key Laboratory of Internet of Things for Smart City, University of Macau, Macao SAR, China. His research interests include intelligent transportation systems, Robotics and computer vision.
\end{IEEEbiography}

\begin{IEEEbiography}[{\includegraphics[width=1in,height=1.25in,clip,keepaspectratio]{{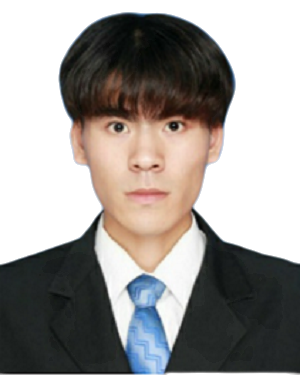}}}]{Fang Li} received the B.S. degree from Harbin Institute of Technology, Weihai, China, in 2020. His research focuses on LiDAR-based 3D object detection during the graduate study and received M.S. degree in mechanical engineering in Beijing Institute of Technology, Beijing, China. His research interests include 3D Object Detection and applications in Autonomous Driving. 
\end{IEEEbiography}

\begin{IEEEbiography}[{\includegraphics[width=1in,height=1.25in,clip,keepaspectratio]{{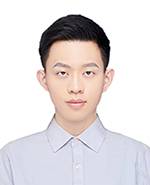}}}]{Peixiang Huang} received his B.S. degree in intelligence science and technology from the School of Intelligent Systems Engineering in Sun Yat-sen University, and 
his M.S. degree in mechanical engineering from the College of Engineering in Peking University. 
His research focuses on intelligent transport systems, camera-based 3D object detection and computer vision.
\end{IEEEbiography}

\begin{IEEEbiography}[{\includegraphics[width=1in,height=1.25in,clip,keepaspectratio]{{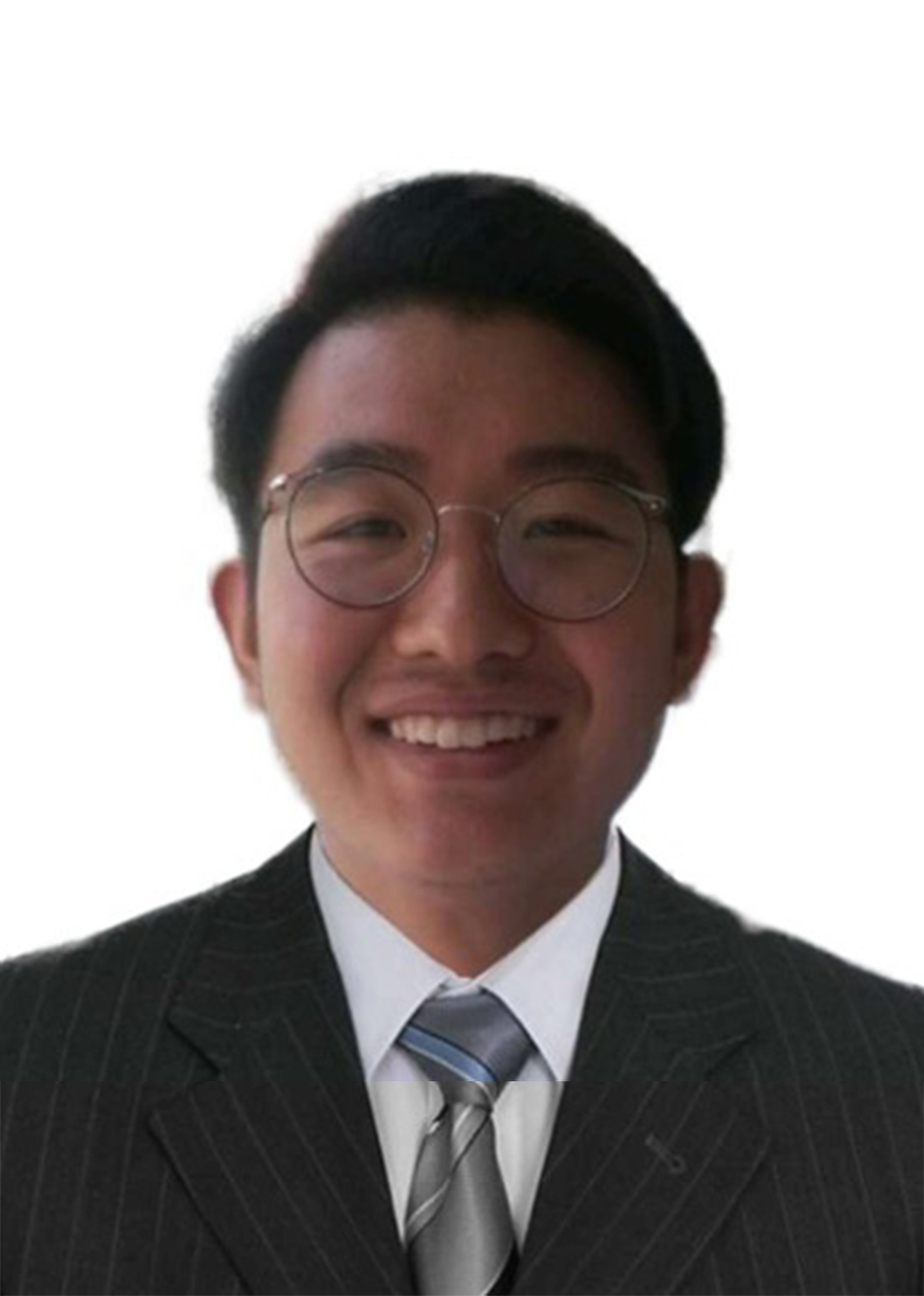}}}]{Ziying Song} was born in Xingtai, Hebei Province, China, in 1997. He received his B.S. degree from Hebei Normal University of Science and Technology (China) in 2019. He received a master's degree from Hebei University of Science and Technology (China) in 2022. He is now a Ph.D. student majoring in Computer Science and Technology at Beijing Jiaotong University (China), with research focus on Computer Vision. 
\end{IEEEbiography}

\begin{IEEEbiography}[{\includegraphics[width=1in,height=1.25in,clip,keepaspectratio]{{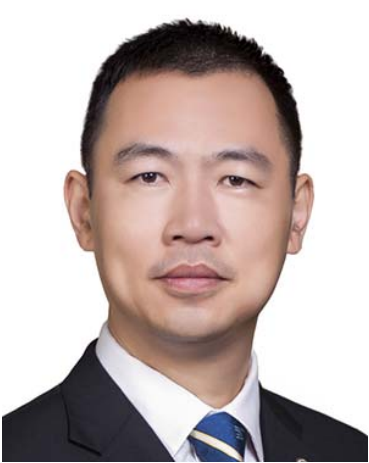}}}]{Zhi-Xin Yang} (Member, IEEE) received the
B.Eng. degree in mechanical engineering from the Huazhong University of Science and Technology, and the Ph.D. degree in industrial engineering and engineering management from the Hong Kong University of Science and Technology, respectively. He is currently an Associate Professor with the University of Macau. His current research interests include robotics, machine vision, intelligent fault diagnosis and safety monitoring.
\end{IEEEbiography}

\end{document}